\documentclass{article}

\usepackage{arxiv}

\usepackage[utf8]{inputenc} 
\usepackage[T1]{fontenc}    
\usepackage{hyperref}       
\usepackage{url}            
\usepackage{booktabs}       
\usepackage{amsfonts}       
\usepackage{nicefrac}       
\usepackage{microtype}      
\usepackage{lipsum}
\usepackage{graphicx}
\usepackage{algorithm, algpseudocode}
\usepackage{amsmath,amsthm,amsfonts,amssymb,amscd, fancyhdr, color, comment, graphicx, environ}
\usepackage{tabularx}
\usepackage{caption}
\graphicspath{ {./images/} }

\title{Semantically-Aware Game Image Quality Assessment}

\author{
 Kai Zhu \\
  \texttt{Kai.Zhu@amd.com} \\
   \And
 Vignesh Edithal \\
  \texttt{Vignesh.Edithal@amd.com} \\
   \And
 Le Zhang \\
  \texttt{Le.Zhang@amd.com} \\
  \And
 Ilia Blank \\
  \texttt{Ilia.Blank@amd.com} \\
  \And
 Imran Junejo \\
  \texttt{Imran Junejo@amd.com} \\
}

\begin{document}
\maketitle
\begin{abstract}
Assessing the visual quality of video game graphics presents unique challenges due to the absence of
reference images and the distinct types of distortions, such as aliasing, texture blur, and geometry
level of detail (LOD) issues, which differ from those in natural images or user-generated content.
Existing no-reference image and video quality assessment (NR-IQA/VQA) methods fail to generalize
to gaming environments as they are primarily designed for distortions like compression artifacts.
This study introduces a semantically-aware NR-IQA model tailored to gaming. The model employs a
knowledge-distilled Game distortion feature extractor (GDFE) to detect and quantify game-specific
distortions, while integrating semantic gating via CLIP embeddings to dynamically weight feature
importance based on scene content. Training on gameplay data recorded across graphical quality
presets enables the model to produce quality scores that align with human perception.
Our results demonstrate that the GDFE, trained through knowledge distillation from binary classifiers,
generalizes effectively to intermediate distortion levels unseen during training. Semantic gating
further improves contextual relevance and reduces prediction variance. In the absence of in-domain
NR-IQA baselines, our model outperforms out-of-domain methods and exhibits robust, monotonic
quality trends across unseen games in the same genre. This work establishes a foundation for
automated graphical quality assessment in gaming, advancing NR-IQA methods in this domain.
\end{abstract}


\section{Introduction}
\begin{figure}
    \centering
    \includegraphics[width=0.6\linewidth]{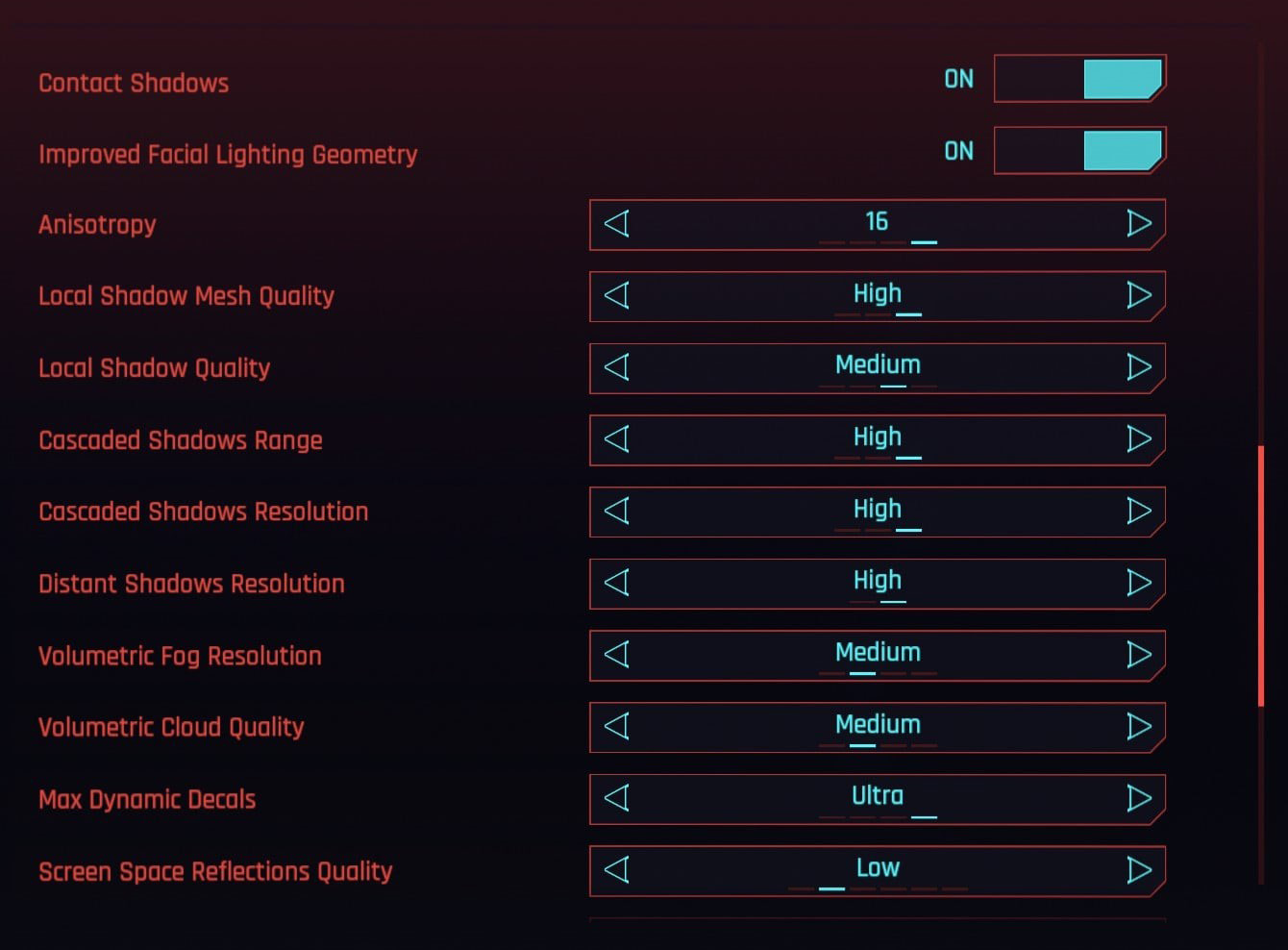}
    \caption{An section of the extensive graphical settings menu in Cyberpunk 2077, illustrating the complexities of
configuration options across multiple pages.}
    \label{fig:1}
\end{figure}

PC gaming is characterized by a vast diversity of hardware configurations, with systems differing in CPU, GPU, and
memory capacities. This variability directly impacts performance, as different setups deliver varying levels of graphical
fidelity and frame rates. Modern games exacerbate this challenge with complex graphical settings, such as texture
quality, ray tracing, level-of-detail (LOD), and shadow details. For example, Cyberpunk 2077 features an extensive
graphical settings menu that spans multiple pages, as shown in Figure \ref{fig:1}.

While experienced gamers may be familiar with these options, determining their specific impact on both performance
and quality for a given hardware setup often involves trial-and-error experimentation. For novice users, the abundance
of technical settings can be overwhelming, leading to suboptimal configurations that fail to fully utilize their hardware’s
potential or deliver the best visual experience.

An automated solution that evaluates graphical settings and dynamically adjusts configurations based on system
capabilities and user preferences would provide significant benefits. Image quality assessment (IQA) is a critical
component of such a solution, offering objective feedback on the impact of settings on the perceived graphical quality.
However, existing IQA and video quality assessment (VQA) methods face two main drawbacks when applied to gaming
environments:

1. Traditional quality metrics such as mean squared error (MSE), peak signal-to-noise ratio (PSNR), and
feature similarity index (FSIM) rely on reference-based approaches that compare a degraded image against
a pristine reference. In the context of gaming, reference images or videos are typically unavailable, as
each gameplay experience is unique, and capturing a reference for every possible scene or configuration is
impractical. This makes reference-based methods unsuitable for real-time assessment of game graphics.

2. Existing no-reference (NR) IQA/VQA methods, including those leveraging deep learning, have predominantly
focused on content such as natural photography, video streaming, or user-generated content (UGC). state-of-
the-art UGC-VQA models like universal video quality (UVQ) \cite{wang2021rich} and comprehensive video quality
evaluator (COVER) \cite{he2024cover} employ convolutional neural networks (CNNs) and vision transformers (ViTs) to
evaluate content and aesthetic qualities, targeting distortions such as compression artifacts, noise, and motion
blur. However, these techniques perform poorly on video game graphics, which exhibit unique distortions like
aliasing, texture blur, low mesh fidelity, and artifacts from specific graphical settings – distortions not aligned with patterns these models are trained to recognize. Consequently, existing NR methods fail to accurately
assess the visual quality of game graphics, resulting in suboptimal performance on gaming content.

The primary goal of this research is to develop a no-reference image quality assessment (NR-IQA) model tailored
specifically for gaming environments, a previously unexplored domain. The model aims to objectively evaluate the
visual quality of game graphics by detecting and quantifying game-specific distortions without relying on reference
images. By processing gameplay images, the model outputs a numerical quality score that reflects the perceived visual
quality from a user’s perspective.

\section{Literature Survey}

In the context of evaluating visual quality, IQA and VQA are closely related tasks. Both aim to quantify perceived
visual quality, and many techniques used in IQA can be extended to VQA by considering temporal information. Since
our work involves assessing the quality of both static images and dynamic content in gaming environments, we discuss
IQA and VQA methods interchangeably.

\subsection{Full reference IQA (FR-IQA)}

Traditional IQA methods rely heavily on reference images to evaluate the quality of distorted images. Common metrics
include peak signal-to-noise ratio (\textbf{PSNR}) and structural similarity index (\textbf{SSIM}). However, these methods often fail
to align with human perception of quality, especially in dynamic and complex visual content \cite{mittal2012no}.

Video multi-method assessment fusion (\textbf{VMAF}) \cite{li2016vmaf}, developed by Netflix, is a full-reference metric that combines
multiple quality assessment algorithms using machine learning to better predict perceived video quality. VMAF
integrates features like visual information fidelity (VIF), Detail loss metric (DLM), and mean co-located pixel
difference (MCPD), and fuses them using SVM-based regression. It has been shown to outperform traditional metrics
such as PSNR and SSIM, making it highly effective for video streaming applications where maintaining consistent
quality across various devices and network conditions is crucial.

FR-IQA methods are impractical for real-time game graphics assessment because pristine reference images are
unavailable during gameplay.

\subsection{CNN-Based NR-VQA}

CNNs have been a popular choice for IQA and VQA due to their ability to extract hierarchical features from images.
Models like ResNet and EfficientNet have demonstrated significant success in various computer vision tasks, including
IQA. CNNs can capture spatial dependencies and complex patterns in visual data, making them well-suited for
identifying both low-level distortions and high-level semantic features \cite{montesinos2022convolutional}. \textbf{UVQ} is a CNN-based model designed to
assess video quality in scenarios where references are unavailable, which is often the case for user generated content
(UGC). The model consists of 3 EfficientNet-based components: ContentNet, DistortionNet, and CompressionNet,
each responsible for evaluating semantic content, technical distortions, and compression artifacts, respectively. The
backbones employ pretrained weights from ImageNet and were finetuned on annotated synthetic distortion datasets [6].
An additional network called AggregationNet spatially and temporally aggregates the features from the sub-networks to
output a single quality score.

In the absence of an existing game graphics IQA model, UVQ seems a reasonable baseline as the UGC training
set contains gameplay footage. However, we demonstrate that its focus on streaming-related distortions renders it
ineffective for game graphical quality assessment.

\subsection{Multi-Modal Methods}

\begin{figure}
    \centering
    \includegraphics[width=0.8\linewidth]{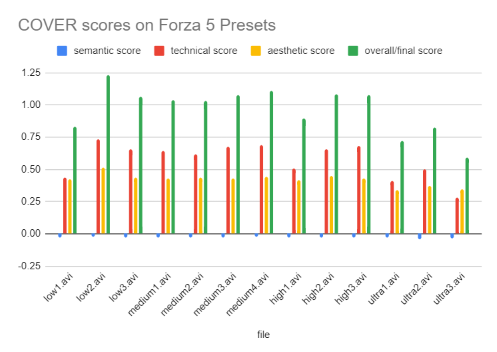}
    \caption{Predicted scores from each of COVER’s semantic, technical, and aesthetic branches, as well as the aggregated
overall score, for a video dataset collected at low, medium, high, and ultra qualities in Forza Horizon 5. The predicted
scores show no correlation with the quality presets (SRCC < 0.2). Ideally, the predicted scores should increase
monotonically from low to ultra qualities. The lack of correlation indicates that COVER is ineffective in the gaming
domain.}
    \label{fig:2}
\end{figure}

Recent advancements in NR-IQA include the use of vision transformers and multi-modal techniques like the contrastive
language–image pre-training (CLIP) model \cite{radford2021learning}, which combines visual and textual information, offering improved
contextual understanding and feature extraction, as demonstrated by \textbf{CLIP-IQA} \cite{wang2023exploring}. disentangled objective video
quality evaluator (\textbf{DOVER}) \cite{wu2023exploring} utilizes a Swin transformer backbone for technical quality and a ConvNet for aesthetic
quality. \textbf{COVER} \cite{he2024cover} further builds on the foundation of DOVER, incorporating an additional semantic branch using a
pretrained CLIP image encoder. By employing a simplified cross-gating block for feature fusion, COVER holistically
assesses video quality and achieves superior performance on multiple UGC video quality datasets in real-time.
While COVER and DOVER performed well on UGC datasets, their focus on aesthetic layout and semantic content
interestingness does not apply to game graphical assessment, where the desired output should be solely based on
graphical quality. Figure \ref{fig:2} demonstrates that applying COVER to the FH5 preset validation data shows no correlation
between preset quality and predicted score.

\section{Methodologies}

\subsection{Challenges and Approach}

\subsubsection{Absence of Human-Annotated Mean Opinion Scores}

Human-labeled mean opinion scores (MOS), commonly used in IQA/VQA training, were unavailable due to resource
constraints. Without MOS, it becomes challenging to train a model that accurately reflects human perception of visual
quality.

To address this, we used game presets (e.g., low, medium, high, ultra) as proxies for quality scores, under the assumption
that game developers configure these presets to reflect overall visual quality. By mapping each preset to a numerical
quality score, we trained the predict quality levels that align with the developers’ intended visual fidelity. This approach
provides a feasible alternative to collecting extensive human-labeled datasets.

However, this method has a significant limitation: lack of granularity. Specifically, there are no ground truth quality
scores for configurations between the preset levels (e.g., an ultra preset with texture quality set to low). This limitation
makes the problem ill-posed and may affect the model’s ability to accurately assess configurations not represented by
the standard presets.

\subsubsection{Limitations of Baseline}

\begin{figure}
    \centering
    \includegraphics[width=1.0\linewidth]{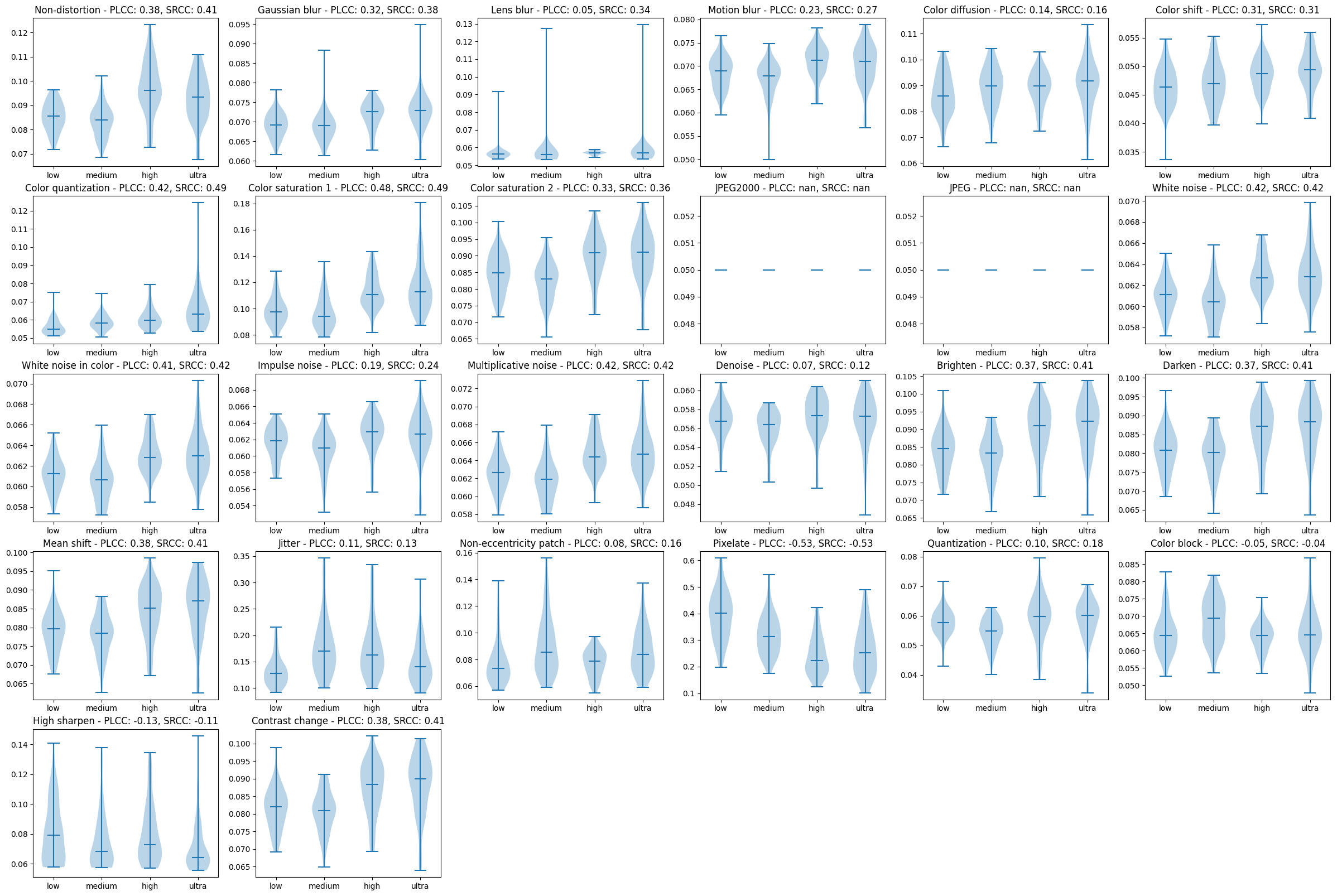}
    \caption{Predicted distortion scores for Forza Horizon 5 validation video data for UVQ’s 25 distortion types, along
with an additional class for non-distortion. No monotonic trend is observable across the quality levels for each of the
distortions, corroborated by the low correlation scores. JPEG compression evaluations output 0 across all quality levels,
since the videos are evaluated at raw capture quality with no compression.}
    \label{fig:3}
\end{figure}

Preliminary experimentation with a regressor trained on UVQ’s DistortionNet demonstrated some correlation. However,
further analysis revealed that DistortionNet’s specific distortion detections exhibited weak or no correlation with
in-game quality settings (Figure \ref{fig:3}). This led to severe overfitting and poor generalization of the regressor based on
DistortionNet features. Consequently, the initial proposal of combining DistortionNet with GDFE was discarded.
Instead, focus shifted to developing the GDFE to detect game-specific distortions more effectively.

\subsubsection{Generalization Issues in Multi-Label Classification from Extreme Data Training}

Due to limited data, we trained each distortion type only on extreme settings (lowest and highest values). Individual
binary classifiers trained on these extremes generalized well to unseen intermediate distortions; their logits correlated
with distortion severity.

However, the multi-label GDFE, trained on the same extremes, overfitted and performed poorly on test data with
multiple distortions, failing to generalize to intermediate settings. Its logits did not correlate with distortion severity, and
performance degraded proportionally to the number of output heads corresponding to the types of detectable distortions.
We hypothesized that the multi-label classification task introduces additional complexity due to the need to model
correlations between different distortions. This increased complexity, combined with limited training data covering
only extreme cases, makes it challenging for the multi-label model to learn decision boundaries that generalize well to
unseen combinations of distortions.

\begin{figure}
    \centering
    \includegraphics[width=1.0\linewidth]{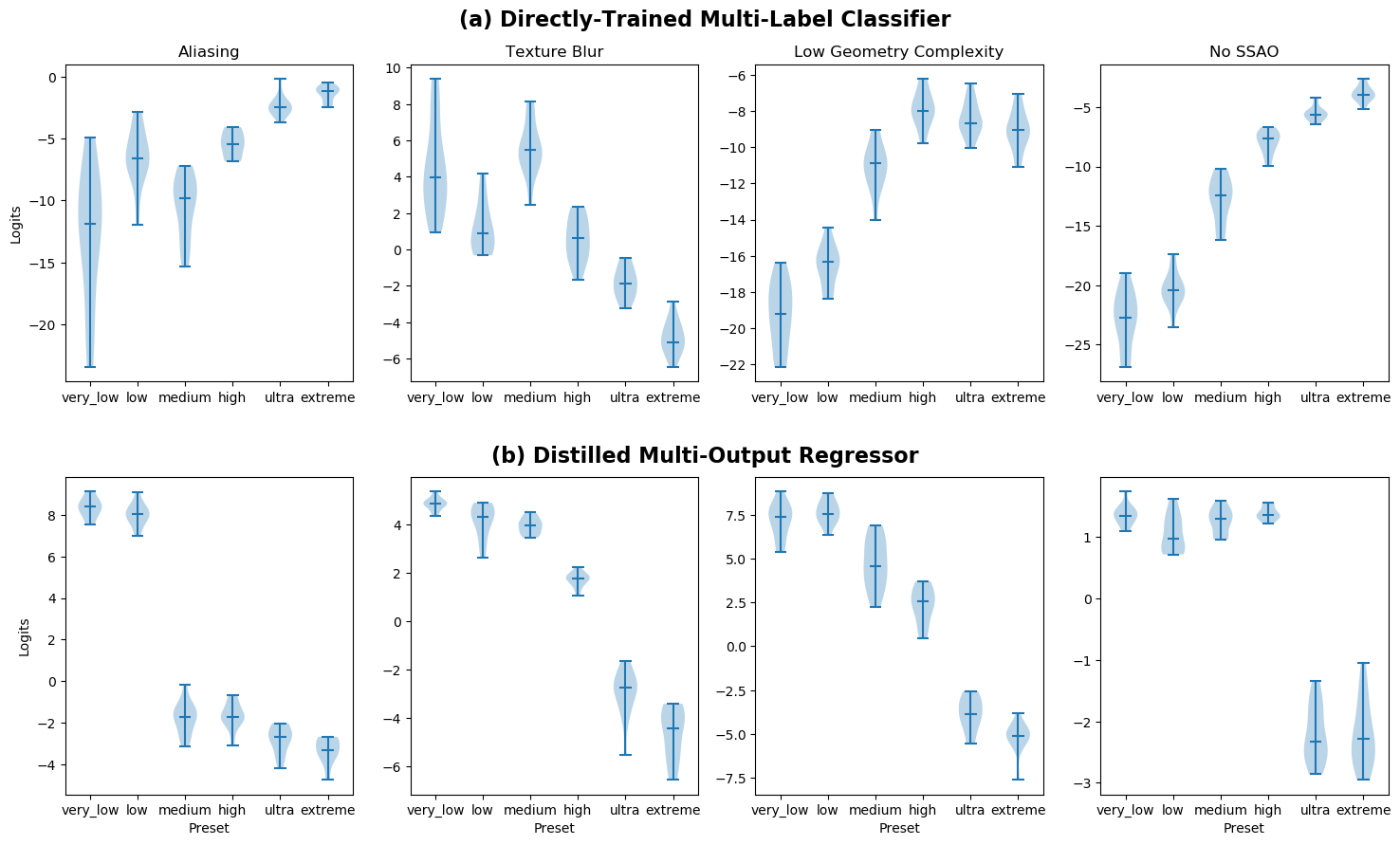}
    \caption{Comparison of the directly-trained GDFE classifier (Model a) and the knowledge-distilled GDFE regressor
(Model b) performance on FH5 test data. In this comparison, higher distortion severity values correspond to lower
visual quality. Model (a) fails to predict a significant correlation between the input features and visual quality, whereas
Model (b) successfully predicts an inverse correlation between distortion severity and quality. Section 5.1 provides a
detailed discussion of these trends.}
    \label{fig:4}
\end{figure}

To address this, we leveraged the binary classifiers’ generalizability by using their logits to pseudo-label both the
distortion and preset datasets, which includes intermediate settings. Through knowledge distillation, we trained
the multi-label GDFE as a regressor on these soft labels, transferring the binary classifiers’ ability to generalize.
Validation confirmed that this approach enabled the GDFE to accurately evaluate unseen intermediate distortion levels.
A comparison of performance of the directly-trained multi-label classifier versus the distilled model on FH5 test set is
shown in Figure \ref{fig:4}.

\subsubsection{Ambiguity in Distortion-Based Quality Regression}

While the GDFE effectively detects game-specific distortions, relying solely on these detections for quality regression
proved ambiguous. For instance, shadow quality detection at night is unreliable because shadows are inherently less
visible or absent, regardless of in-game settings. Similarly, the visibility of ambient occlusion depends on the content
of the scene, such as the presence of buildings or objects that cast shadows. These factors introduce variability that
reduces the reliability of quality regression based on distortion features.

Drawing inspirations from COVER’s semantically-gated technical evaluation, we incorporated image semantics into
the evaluation of distortions in the quality regression network. The goal is to leverage semantic information to better
contextualize the distortions based on scene content. This semantic gating allows the model to discern when certain
distortions are contextually appropriate or when their detection should influence the quality score.

\section{Proposed Method}

\begin{figure}
    \centering
    \includegraphics[width=1.0\linewidth]{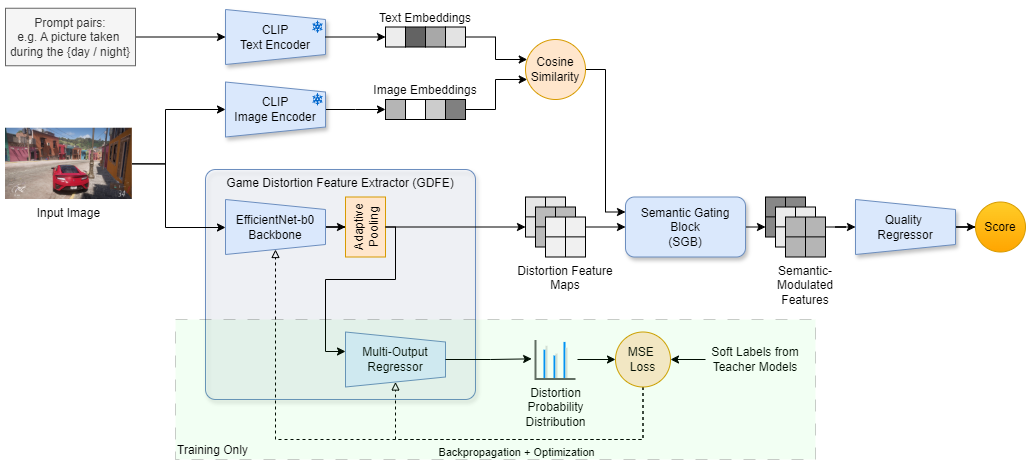}
    \caption{An overview of the game quality assessment model, featuring the game distortion feature extractor (GDFE)
trained via knowledge distillation using soft labels from teacher binary classifiers. The GDFE’s feature maps are gated
using image-text similarity to produce semantic-aware quality scores.}
    \label{fig:5}
\end{figure}

The overall architecture of our model is illustrated in Figure \ref{fig:5}. The core components include the game distortion
feature extractor (GDFE), pretrained CLIP text and image encoders, and a quality regressor. The GDFE detects
gaming-specific distortions, and the CLIP model provides semantic information about the image content. By combining
features from both through a semantic gating block (SGB), the quality regressor produces a robust visual quality score.

\subsection{GameVQA Datasets}

The scarcity of annotated game video datasets poses a significant challenge for training VQA models for gameplay.
Existing datasets, such as GamingVideoSet \cite{barman2018gamingvideoset}, focus on distortions relevant only to video streaming such as the
effects of video compression encodings.

To address this gap, we collected gameplay video from Forza Horizon 5 (FH5) under diverse conditions, including
different vehicle, biome, and times of day. We specifically chose FH5 for data collection because the racing genre offers
reproducibility through consistently defined tracks and environmental conditions. This reproducibility allows us to
minimize inconsistencies unrelated to our features of interest, which is crucial when working with a smaller dataset
that lacks the averaging effect of large, diverse data. We utilized Autobot, an in-house automatic gameplay agent, to
configure and play the game, while OBS Studio was used recording.

Videos were recorded in RAW format without encoding at 1080p resolution and a sampling rate of 1 frame per second
(FPS). To minimize class imbalance bias, we ensured a similar number of frames were recorded for each data category. We created two separate datasets to train the game distortion feature extractor (GDFE) and the quality regressor, respectively.

\subsubsection{Distortion Dataset}

The FH5 distortion dataset, used to train the GDFE, initially consisted of approximately 4,000 frames (67
minutes of video) for each of the four distortion classes and the undistorted class. The undistorted videos were collected
using the "Extreme" preset. The distorted data were collected using the same preset with specific modifications to the
graphical settings:

\begin{itemize}
  \item \textbf{Aliasing}: anti-aliasing turned off.
  \item \textbf{Texture Blur}: Texture quality set to low.
  \item \textbf{Low mesh LOD}: Mesh quality set to very low.
  \item \textbf{SSAO Off}: Screen-space ambient occlusion (SSAO) set to off.
\end{itemize}

These specific graphical settings were selected to provide diversity in detection complexity. For example, aliasing
represents a low-level distortion localized on object outlines, while SSAO affects higher-level features near concave
intersections of 3D geometries.

Additionally, the distortion dataset was supplemented with shadow quality data. Unlike other settings, shadow
configurations include an "Off" setting in addition to ranging from "Low" to "Ultra." The visual difference between the
"Off" setting and the others is significantly greater than between the non-off settings. To account for this, and leveraging
the knowledge distillation pipeline described in Section 4.2, shadow quality data were collected at "Off," "Low," and
"High" settings to provide training at different distortion severities, adding a total of 8,900 frames.

\begin{figure}
    \centering
    \includegraphics[width=1.0\linewidth]{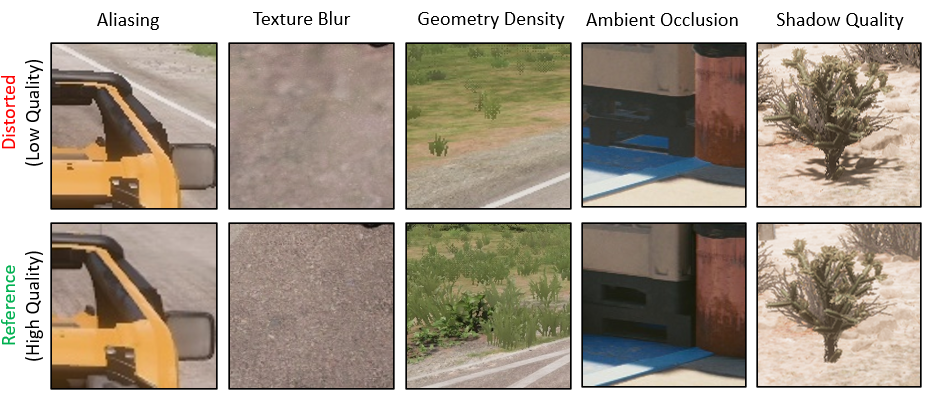}
    \caption{The five quality indicators in the distortion dataset. \textbf{Aliasing} appears as jagged or stair-stepped outlines.
\textbf{Texture blur} indicates low-resolution textures. \textbf{Geometry density} reflects low levels of detail in 3D objects. \textbf{Ambient
occlusion} is evident in contact shadows, the absence of which makes objects appear to float. \textbf{Shadow quality} refers to
missing shadows or shadows rendered at low resolution.}
    \label{fig:6}
\end{figure}

Figure \ref{fig:6} shows examples of each type of distortion included in this dataset. A test set containing a total of 1,446 frames was created, with approximately equal distribution among the distortion
types.

\subsubsection{Preset Dataset}

To train the quality regressor using preset data as proxies for ground truth scores, we recorded approximately 70 minutes
of video content for each graphical preset—very low, low, medium, high, ultra, and extreme—for a total of 25,000
frames. Since the anti-aliasing setting remains unchanged across these presets, we assigned specific multi-sampling
anti-aliasing (MSAA) settings to each quality preset, partially based on the corresponding presets in Forza Horizon 4:

\begin{itemize}
    \item \textbf{Very Low Preset}: MSAA set to off.
    \item \textbf{Low Preset}: MSAA set to off.
    \item \textbf{Medium Preset}: MSAA set to 2×.
    \item \textbf{High Preset}: MSAA set to 2×.
    \item \textbf{Ultra Preset}: MSAA set to 2×.
    \item \textbf{Extreme Preset}: MSAA set to 8×.
\end{itemize}

The test set comprises 11,000 frames, including 14 minutes of footage for each of the six presets and seven additional
configurations to evaluate the model’s granularity in detecting quality changes between presets. These additional
configurations are:

\begin{itemize}
    \item Low preset with 2× MSAA
    \item Ultra preset with SSAO off
    \item Extreme preset with SSAO off
    \item Extreme preset with MSAA off
    \item Extreme preset with both MSAA and SSAO off
    \item Extreme preset with high texture quality
    \item Extreme preset with 2× MSAA
\end{itemize}

\begin{figure}
    \centering
    \includegraphics[width=0.5\linewidth]{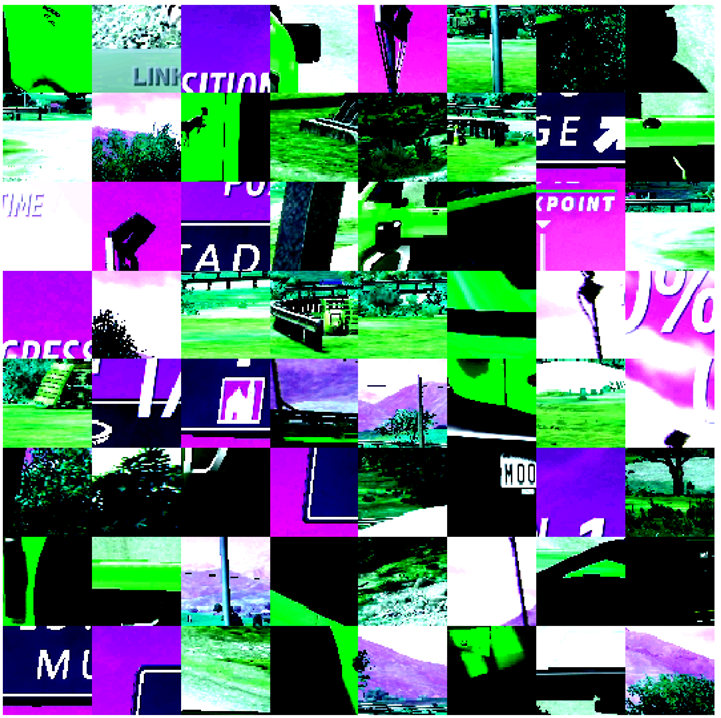}
    \caption{Normalized and augmented training image for the aliasing detector, utilizing an edge-focused fragmentation
strategy. While patch sampling may occasionally capture UI elements like text, most patches include real rendered
outlines that, in this example, exhibit aliasing artifacts.}
    \label{fig:7}
\end{figure}

\subsection{Game distortion feature extractor (GDFE) Knowledge Distillation}

To effectively detect and quantify the unique distortions present in game graphics, we developed the game distortion
feature extractor (GDFE). Initial attempts to train the GDFE directly as a multi-label classifier proved ineffective,
especially as the number of distortion types increased (as discussed in 3.1.3). To address this, we adopted a knowledge
distillation approach, where a set of specialized teacher models informs the training of a unified student GDFE model
as a multi-output regressor.

The following sections detail the teacher models and the student model, including architectural choices, training
strategies, and optimizations employed.

\subsubsection{Teacher Models}

The teacher models are individual binary classifiers, each dedicated to detecting a specific type of distortion. This modular
approach allows for flexibility and fine-tuning tailored to the characteristics of each distortion. After experimenting
with various architectures such as ResNet18, ResNet50, and Swin transformer, we selected EfficientNet-B0 \cite{tan2019efficientnet} as
the backbone for the teacher models due to its balance of performance and computational efficiency. Initialized with
weights pretrained on the ImageNet dataset, the backbone features are globally max-pooled and passed through a simple
head consisting of a randomly-initialized linear layer with an output size of one.

Each binary classifier, unless otherwise specified, was finetuned on its respective distortion dataset using the AdamW
optimizer with a learning rate of $1 * 10^{-3}$, Binary cross entropy (BCE) loss, and a batch size of 48. To mitigate
overfitting, we adopted an early stopping strategy, applied a weight decay of $1 * 10^{-2}$, and added a dropout layer in the
head with a dropout probability of 0.2. Data augmentation techniques—including horizontal and vertical flips, and
color jitter (brightness: 0.1, contrast: 0.1, saturation: 0.1, hue: 0.2)—were employed to increase training diversity.

\textbf{Aliasing Detection}: For aliasing detection, we employed a fragmentation strategy inspired by DOVER \cite{wang2023exploring} to focus on
technical distortions and reduce reliance on semantic features. Specifically, input images were cropped into 64 × 64
pixel patches. To emphasize edge information crucial for aliasing detection, we applied a Sobel filter to each patch
to detect edges and calculated the standard deviation of the Sobel-filtered map to quantify edge content. We then
sampled 64 patches with probabilities proportional to their edge content and reassembled them into 512 × 512 images
for training (Figure \ref{fig:7}).

\textbf{SSAO Detection}: To improve generalizability for SSAO detection, we augmented the training data with 1,000 images
from Cyberpunk 2077, in addition to the FH5 distortions. We utilized BCE with Variance Loss to enhance detection
consistency:

$$ \mathcal{L} = \mathcal{L}_{\mathrm{BCE}} + \lambda \cdot \frac{1}{C} \sum_{c=1}^{C} \mathrm{Var}(\mathbf{z}_c) $$

\begin{figure}
    \centering
    \includegraphics[width=1.0\linewidth]{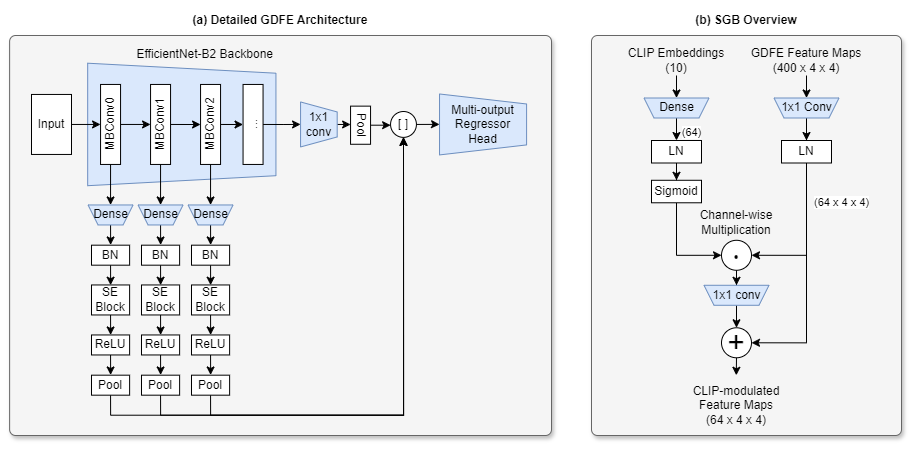}
    \caption{\textbf{(a)} The modified backbone of the game distortion feature extractor (GDFE) incorporates skip connections
from the first three MBConv blocks. These connections are processed through batch normalization (BN), squeeze-andexcitation
(SE) blocks, and adaptive max pooling to rescale features to 4 × 4. The rescaled features are concatenated
with the backbone output to provide multi-scale features for distortion detection. \textbf{(b)} The Semantic Gating Block
(SGB) projects CLIP similarity vectors to match the compressed channel size of the GDFE feature maps, followed by
channel-wise multiplication to gate the feature maps using semantic information.}
    \label{fig:8}
\end{figure}

where $C$ is the number of distortion classes, $Z_c$ represents the predicted logits for class $c$, and $ \lambda $ is a hyperparameter
controlling the variance penalty. This loss function penalizes high variance in predictions across similar inputs,
encouraging consistent outputs.

\textbf{Shadow Quality Detection}: Shadow quality detection was approached as a regression task due to the non-linear
perceptibility of shadow distortions. We assigned numerical values to the distortion levels: "Off" (-1), "Low" (-0.6),
"High" (-0.2), and "Ultra" (1). This allows the model to approximate the relationship between shadow quality settings
and perceived distortion severity. The model is trained using mean squared error (MSE) loss.
The teacher models generated soft labels for their respective distortions on the distortion dataset and on a portion
(one-third) of the preset training set, providing rich information for the student model.

\subsubsection{Student Model}

The GDFE student model was trained as a multi-output regressor using knowledge distillation from the teacher models.
By leveraging the soft labels provided by the teachers, the student model learns to predict continuous values representing
the severity of each distortion.

To enhance detection capacity for multiple features, we selected EfficientNet-B2 \cite{tan2019efficientnet} as the backbone, offering
increased representational power. To improve detection of distortions at various scales, especially lower-level features,
we modified the backbone by incorporating skip connections from the first three MBConv layers (Figure \ref{fig:8}a). These
skip connections passed feature maps through squeeze-and-excitation blocks, projected and pooled them to 4 × 4,
and concatenated them to form a final feature map with 400 channels, inspired by UVQ’s DistortionNet \cite{wang2021rich}. The
regressor head outputs five values corresponding to the five distortion types and is used only during training to facilitate
meaningful feature extraction.

The student model was trained using MSE loss on the pseudo-labeled data. Optimization was performed with the
AdamW optimizer using a learning rate of $1 * 10^{-4}$, weight decay of $1 * 10 ^ {-4}$, and a dropout rate of 0.2. Data
augmentation techniques identical to those used in the teacher models were applied to diversify training images. This knowledge distillation approach enables the student model to generalize effectively to intermediate distortion
levels and handle multiple simultaneous distortions.

\subsubsection{CLIP Encoders}

We employ pretrained and frozen CLIP encoders to extract semantic information from images. The CLIP image encoder
generates image embeddings, while the CLIP text encoder processes textual prompts. Specifically, we use 10 pairs of
positive and negative prompts to capture various aspects of image content, listed in Table \ref{table:1}.

\renewcommand{\arraystretch}{1.5} 

\captionsetup[table]{skip=12pt} 

\begin{table}[h]
\centering
\begin{tabular}{|l|l|}
\hline
\textbf{Positive Prompts} & \textbf{Negative Prompts}  \\
\hline
A picture taken during the daytime & A picture taken at night \\
A picture with a car & A picture with no cars \\
A picture of a building & A picture with no buildings \\
A picture with grass & A picture with no grass \\
A picture with trees & A picture with no trees \\
A picture of a city & A picture of wilderness \\
A picture with a lot of details & A simple picture with minimal content \\
\hline
\end{tabular}
\caption{Positive and Negative Prompts for CLIP Encoders
}
\label{table:1}
\end{table}

Cosine similarities between the image embeddings and both positive and negative text embeddings are computed.
For each positive-negative pair, the similarities are normalized using a softmax activation function. These scores are
aggregated into a 10-dimensional similarity vector, where each value represents the probability of the image matching a
specific positive prompt. This vector effectively captures the semantic context relevant to quality assessment.

\subsubsection{Semantic gating block (SGB)}

The semantic vectors derived from the CLIP encoders are integrated with the image feature maps from the GDFE
through the Semantic gating block (SGB), adapted from COVER’s simplified cross-gating block (SCGB) \cite{he2024cover}:

Summarized in Figure [\ref{fig:8}]b, the semantic vectors are first projected to match the channel size of the GDFE feature maps.
A sigmoid activation function is then applied to the projected vectors to generate gating coefficients. These coefficients
are multiplied channel-wise with the GDFE feature maps, dynamically modulating the importance of each feature
channel based on the image content.

This gating mechanism allows the model to adjust the weighting of feature channels dynamically, depending on the
semantic context of the image. For instance, in nighttime scenes, the evaluation of shadow quality is deprioritized,
allowing the model to focus on more relevant distortions.

\subsubsection{Quality Regressor}

The quality regressor generates the final visual quality score using the semantically gated feature maps. It is trained
on the preset dataset, where each graphical preset (very low, low, medium, high, ultra, extreme) is mapped to a
corresponding numerical value of 0 to 5.

The quality regressor is trained using MSE loss between the predicted scores and the preset-mapped score values.
However, using quality presets as proxies for MOS introduces challenges:

\begin{itemize}
    \item \textbf{Sparse Ground Truths:} Scores are only available for preset configurations, leaving gaps for in-between
settings. For example, no ground truth exists for the visual quality of the High preset with texture quality set to
low, compared to Medium preset with texture quality set to Ultra.
    \item \textbf{Ill-Posed Problem:} Multiple graphical settings change simultaneously (e.g., aliasing, texture quality), allowing
the model to potentially focus on a single feature while neglecting others, leading to biased assessments.
\end{itemize}

To mitigate overfitting and promote balanced evaluation across all distortion features, we implement the following
strategies:

\begin{itemize}
    \item \textbf{Weight Decay:} Applied to discourage the model from assigning excessive weights to specific features,
promoting uniform importance across all features.
    \item \textbf{Dropout:} Incorporated within the regressor to prevent reliance on any single feature, encouraging the use of a
diverse set of features for quality assessment.
\end{itemize}

By integrating regularization techniques and leveraging semantically gated feature maps, the quality regressor improves
its ability to generalize to unseen intermediate settings and assess quality based on multiple simultaneous distortions.

\section{Results and Discussion}

\subsection{Evaluation of GDFE}

Table \ref{table:2} presents an ablation study assessing the impact of skip connections and SE blocks on the GDFE student model.
The architecture incorporating both skip connections from the first three MBConv blocks and SE blocks achieved the
lowest validation loss, highlighting the effectiveness of our modifications.

\begin{table}[h]
\centering
\begin{tabular}{|l|l|}
\hline
\textbf{Model} & \textbf{Validation MSE}  \\
\hline
No skip connections & 5.314 \\
2 skip connections & 4.811 \\
3 skip connections & 4.760 \\
\textbf{3 skip connections + SE} & \textbf{4.665} \\
\hline
\end{tabular}
\caption{Ablation of GDFE student model demonstrating the performance benefits of incorporating skip connections
from the first three MBConv blocks and squeeze-and-excitation (SE) blocks. The selected design achieves lower
validation loss compared to variants with fewer skip connections or without SE blocks.
}
\label{table:2}
\end{table}

\begin{figure}
    \centering
    \includegraphics[width=1.0\linewidth]{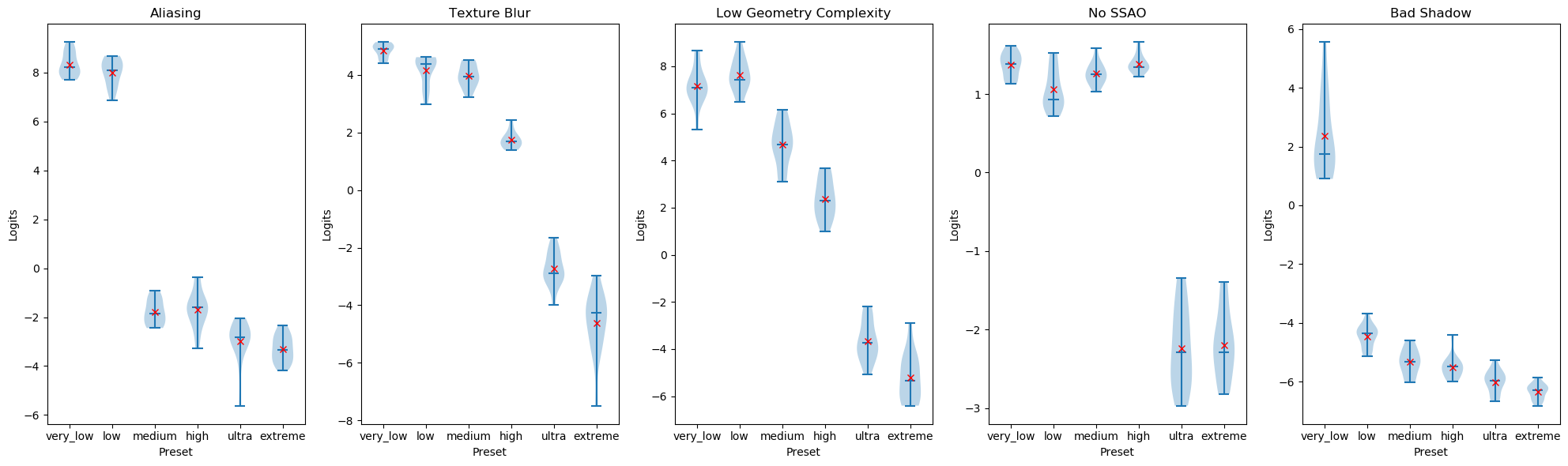}
    \caption{Distributions of GDFE distortion predictions on the FH5 preset test data. Higher values indicate greater
distortion severity and lower visual quality. Horizontal bars represent the median, lower, and upper bounds, while the
red x marks the mean prediction for each preset.}
    \label{fig:9}
\end{figure}

Figure \ref{fig:9} summarizes the GDFE’s distortion predictions on the FH5 preset test dataset. To reduce variability caused by
the stochastic nature of the visual complexity and contents in the scenes, we plot the distributions of mean scores from
groupings of 10 randomly sampled frames.

The aliasing results align with in-game configurations: very\_low and low presets, which have MSAA off, show high
distortion levels, while medium to extreme presets with 2× MSAA demonstrate consistently low distortion levels.
Similarly, SSAO is disabled for very\_low to high presets, and its absence is clearly reflected in the GDFE outputs for
these presets.

For texture, geometry, and shadow settings, distortion predictions generally decrease monotonically as preset quality
improves, reflecting the expected trends based on the graphical configurations. However, geometry LOD distortion
shows a minor anomaly where the very\_low preset exhibits slightly lower distortion than the low preset, indicating
potential challenges in distinguishing fine-grained differences in the low-end qualities for this distortion type.

\begin{figure}
    \centering
    \includegraphics[width=1.0\linewidth]{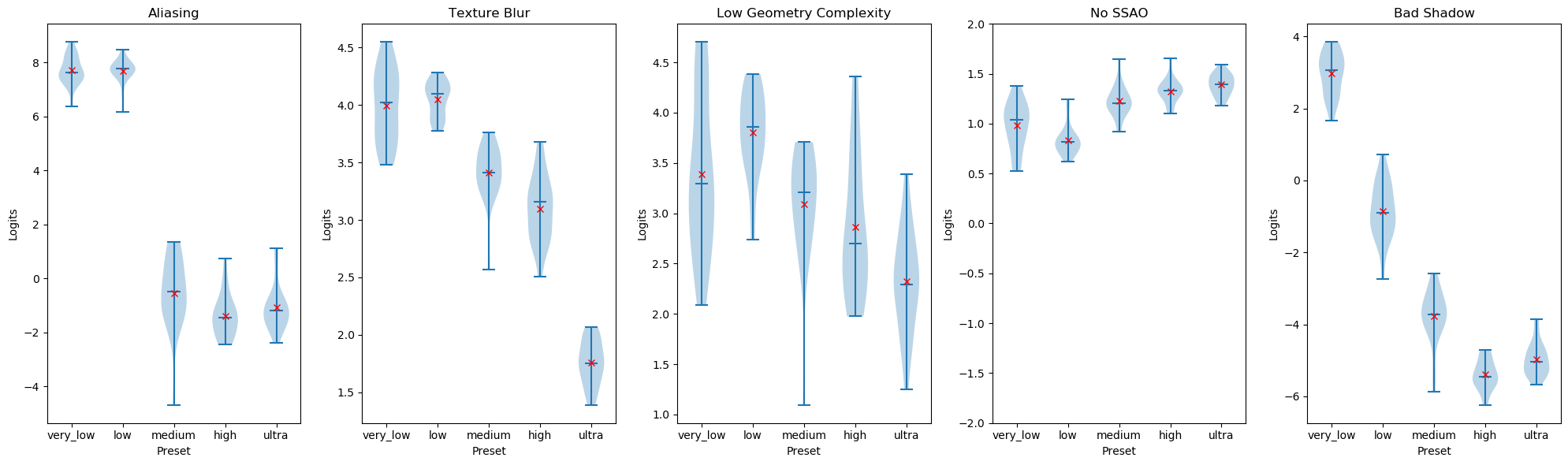}
    \caption{Distributions of GDFE distortion predictions on a Forza horizon 4 (FH4) preset dataset, showcasing
generalizability on an unseen game in the same genre.}
    \label{fig:10}
\end{figure}

To further evaluate the robustness of GDFE, we tested it on a Forza horizon 4 (FH4) preset dataset. As shown in Figure
\ref{fig:10}, the predictions for aliasing, texture, and shadow distortions align closely with the expected trends observed for FH5.
Geometry detection, however, reproduces the same anomaly on the very\_low preset.

Interestingly, GDFE predicts SSAO as disabled across all presets. Visual inspection by a human observer confirmed the
absence of observable SSAO effects, potentially due to an in-game bug that prevents proper rendering of this feature.
These result demonstrate that GDFE, as an objective quality assessment module, effectively generalizes to an unseen
game within the same genre, retaining its predictive performance for most distortion types.

\subsection{Evaluation of Quality Regressor}

\begin{figure}
    \centering
    \includegraphics[width=1.0\linewidth]{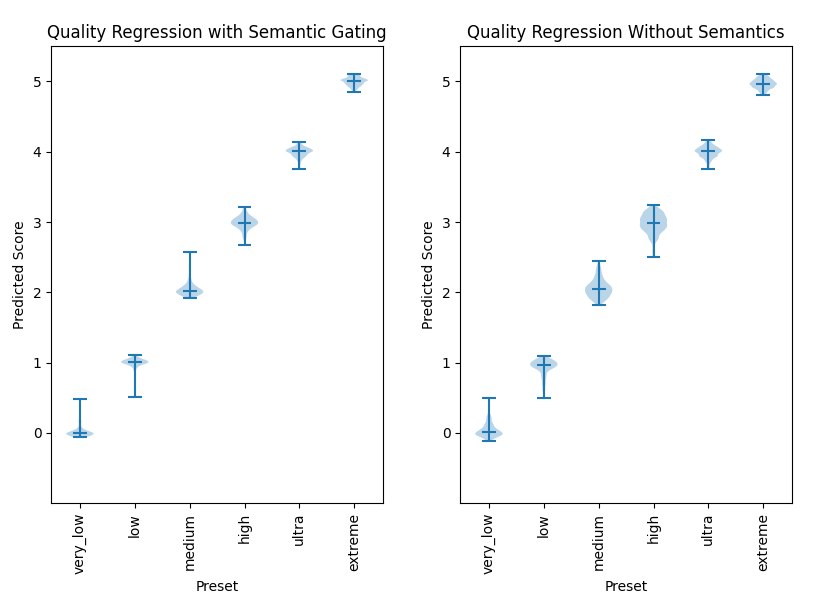}
    \caption{Distribution of quality regressor predictions on FH5 preset test data, comparing output with and without
semantic gating. Quality regression with semantically-gated distortion features produce distributions with lower
variance.}
    \label{fig:11}
\end{figure}

Figure \ref{fig:11} compares the predictions of the quality regressor with and without semantic integration from the SGB
block. Both versions produce distributions for the FH5 test data with medians aligned to the assigned quality scores.
However, incorporating CLIP-encoded semantic embeddings reduces prediction variance, with a mean variance of
0.008 compared to 0.014 for the model without semantic gating. While the overall difference is subtle, it demonstrates
the stabilizing effect of semantic gating.

Both models achieve a Spearman correlation coefficient (SRCC) of 0.986, indicating a strong monotonic correlation
between predicted and true scores. However, it is important to note that the true scores are sparse, limited to integer
values associated with each quality preset. This sparsity likely inflates the SRCC compared to a scenario where
subjective mean opinion scores (MOS) are available as ground truth.

We further evaluated the model’s ability to detect granular quality changes resulting from individual setting adjustments
rather than preset configurations. Figure 12 illustrates how modifying single settings impacts the detected quality score.
For instance, increasing the MSAA setting on the low preset raises the quality score closer to that of the medium preset.
Similarly, reducing texture quality on the extreme preset from extreme to high lowers the quality score, aligning it
closer to the high preset.

A significant limitation of using presets as ground truth is the unconstrained nature of quality changes resulting from
individual setting adjustments. For example, the model predicts a higher quality for the extreme\_no\_aa configuration
compared to extreme\_high\_texture, but there is no ground truth to verify whether this reflects true perceptual quality
differences. Similarly, changes in anti-aliasing (AA) settings show a large impact on predictions, likely due to the
GDFE’s strong confidence in alias distortion detection rather than AA’s actual perceptual importance.

The sparse nature of preset-based truth scores also complicates quantitative evaluation. Without ground truth for
non-preset configurations, metrics such as Spearman correlation coefficient (SRCC) or mean squared error (MSE)
cannot be calculated, limiting formal assessment of the model’s predictions.

To address these challenges, a diverse dataset containing both preset and mixed settings with human-annotated mean
opinion scores (MOS) is needed. Fine-tuning the model on this data would improve alignment with human perception
and enable robust quantitative evaluation.

\begin{figure}
    \centering
    \includegraphics[width=1.0\linewidth]{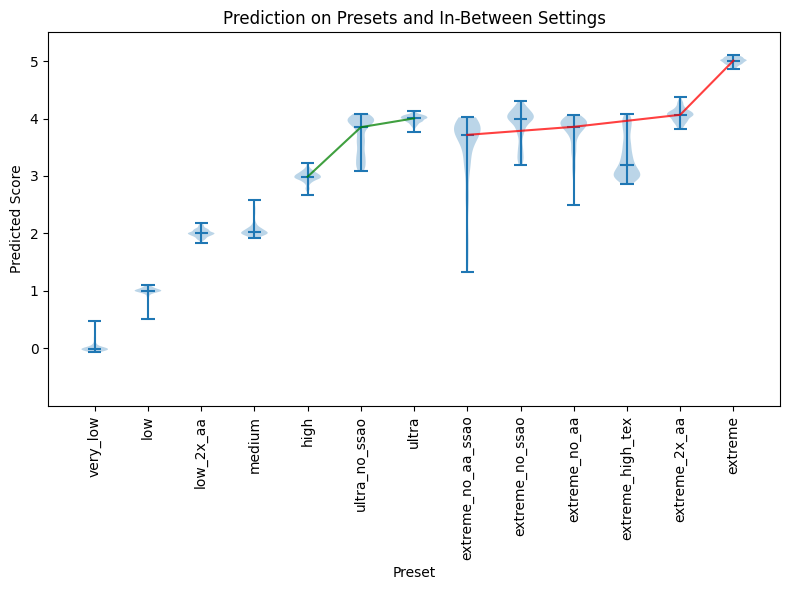}
    \caption{Quality regressor predictions for additional FH5 configurations. The green line shows the modified ultra
preset (SSAO disabled) scoring between the original ultra and high presets. The red line indicates decreasing quality for
the extreme preset as MSAA is reduced from 8× to 2×, then off, and further when both MSAA and SSAO are disabled.}
    \label{fig:12}
\end{figure}

Lastly, we evaluated the model on Forza horizon 4 (FH4) and Forza motorsport (FM), two unseen games within
the same racing genre. Predicted scores show a monotonic increase across presets, with SRCCs of 0.966 and 0.880,
respectively, displaying generalizability that outperforms the previously mentioned out-of-domain UGC models.
Detection granularity is reduced for these unseen titles, particularly for FM, where the model detects only minor
differences between low and medium presets and between high and ultra presets. Interestingly, FM presets consistently
score higher than those of FH4, likely due to FM’s improved graphics, as it was released in 2023 compared to FH4’s
release in 2018.

\section{Conclusion and Future Research Plans}

\subsection{Summary and Key Findings}

This research addressed the primary goal of developing a no-reference image quality assessment (NR-IQA) model
tailored to gaming environments, a domain that presents unique challenges. The proposed model evaluates the visual
quality of game graphics by detecting and quantifying game-specific distortions, such as aliasing, texture blur, and
geometry LOD, without relying on reference images. By leveraging pretrained EfficientNet backbones, a knowledgedistilled
Game distortion feature extractor (GDFE), and semantically-aware gating through CLIP embeddings, the
model integrates perceptual semantics with technical quality evaluation to produce robust and meaningful quality
predictions.

Key findings include:

\begin{itemize}
    \item \textbf{Effectiveness of Binary Classifiers and Knowledge Distillation:} Training individual binary classifiers on
extreme distortion levels effectively captured fine-grained distortion features. Using knowledge distillation to
pseudo-label data and train a student multi-output regressor enabled the GDFE to generalize to intermediate
distortion levels unseen during training, addressing the limitations of direct multi-label classification, which
struggled with overfitting and poor generalization.
    \item \textbf{Semantic Gating for Contextual Awareness:} Incorporating CLIP-encoded semantic embeddings into the
quality assessment pipeline reduced prediction variance and improved robustness by dynamically weighting
feature channels based on image context.
    \item \textbf{Generalizability Across Games:} Although trained on data from a single game (Forza horizon 5), the model
demonstrated the ability to generalize to unseen titles within the same racing genre. Generalizability is reduced
for unseen games, but the observed monotonic trends and meaningful predictions reflect the model’s robustness
despite its limited training scope.
\end{itemize}

\subsection{Significance and Broader Impacts}

The significance of this research extends beyond the immediate goal of video game quality assessment:

\begin{enumerate}
    \item \textbf{Broader Research Applications:} The methodology, particularly the use of knowledge distillation and
semantic gating, is adaptable to other domains requiring no-reference quality assessment such as medical
imaging.
    \item \textbf{Industry Benefits}: This research highlights the potential for an automated graphical settings
recommendation system. Such a system could dynamically optimize game configurations to balance visual
quality and performance, enhancing both user experience and hardware utilization.
    \item \textbf{Academic Contribution:} The findings contribute to the growing field of no-reference quality assessment,
providing a scalable solution to detect and quantify complex, domain-specific distortions.
\end{enumerate}

\subsection{Future Research Directions}

This study opens several avenues for further exploration:

\begin{itemize}
    \item \textbf{Dataset Expansion:} Extending the dataset to include a broader range of games, genres, and graphical
distortions—such as lighting effects and particle rendering—will improve model robustness and applicability.
    \item \textbf{Human-Labeled MOS Collection:} Addressing the limitations of preset-based scores by collecting human-labeled
mean opinion scores (MOS) would enhance the model’s ability to evaluate intermediate configurations
and align more closely with human perception.
    \item \textbf{Optimizing Model Performance:} Applying model compression techniques, such as quantization or pruning,
could significantly reduce computational overhead, enabling real-time quality assessment on more diverse
hardware. This is particularly relevant for CLIP-encoding branch of the model, which represents a significant
performance bottleneck due to its high processing time.
\end{itemize}

\newpage

\bibliographystyle{unsrt}  
\bibliography{main}

\end{document}